\pdfoutput=1

\documentclass[11pt]{article}

\usepackage[final]{acl}

\usepackage{times}
\usepackage{latexsym}

\usepackage[T1]{fontenc}

\usepackage[utf8]{inputenc}

\usepackage{microtype}

\usepackage{inconsolata}

\usepackage{graphicx}
\usepackage{bibentry}
\usepackage{xcolor}
\usepackage{booktabs}
\usepackage{graphics}
\usepackage{booktabs}
\usepackage{geometry}
\usepackage{array}
\usepackage{caption}
\usepackage{comment}
\usepackage{tcolorbox}
\usepackage{makecell}
\usepackage{longtable}

\title{Monolingual and Multilingual Misinformation Detection\\ for Low-Resource Languages: A Comprehensive Survey}

\author{
  \textbf{Xinyu Wang\textsuperscript{1,*}},
  \textbf{Wenbo Zhang\textsuperscript{1,*}},
  \textbf{Sarah Rajtmajer\textsuperscript{1}}
\\
  \textsuperscript{1}College of Information Sciences and Technology, The Pennsylvania State University, USA
\\
  \texttt{
    \{xzw5184, wjz5120, smr48\}@psu.edu
  }
}

\begin{document}
\maketitle
\def\thefootnote{*}\footnotetext{These authors contributed equally to this work.}\def\thefootnote{\arabic{footnote}}
\begin{abstract}
In today's global digital landscape, misinformation transcends linguistic boundaries, posing a significant challenge for moderation systems. 
Most approaches to misinformation detection are monolingual, 
focused on high-resource languages, i.e., a handful of world languages that have benefited from substantial research investment. 
This survey provides a comprehensive overview of the current research on misinformation detection in low-resource languages, both in monolingual and multilingual settings. We review existing datasets, methodologies, and tools used in these domains, identifying key challenges related to: data resources, model development, cultural and linguistic context, and real-world applications. 
We examine emerging approaches, such as language-generalizable models and multi-modal techniques, and emphasize the need for improved data collection practices, interdisciplinary collaboration, and stronger incentives for socially responsible AI research. Our findings underscore the importance of systems capable of addressing misinformation across diverse linguistic and cultural contexts.

\end{abstract}

\section{Introduction}

In today's interconnected digital world, information flows seamlessly across geographical and linguistic boundaries, through online and offline spaces \cite{darvin2016language,gee2011language}. The increased use of local languages among international communities online has led to a multilingual Internet, enabling both local and global participation in information exchange \cite{lee2013language}.
This global digital ecosystem also presents unique challenges, as the proliferation of misinformation transcends any single language or region, manifesting as a global phenomenon that impacts online societies worldwide, as exemplified during the pandemic \cite{shahi2020fakecovid}. In this context, false information can quickly cross linguistic borders, challenging authoritative organizations to respond promptly to the spread of fake news, thus necessitating a move beyond an English-centric view \cite{mohawesh2023multilingual}. As multilingualism permeates social communities, the interweaving of languages generates new hybrid meanings and grammatical structures \cite{darvin2016language}. The blending of diverse languages and modalities in digital texts can lead to both intentional attempts to evade detection and moderation, as well as unintentional creation of unconventional idioms and word choices. These complexities make the accurate identification and control of false information significantly more difficult.

\begin{figure}[ht]
    \centering
    \includegraphics[scale=0.36]{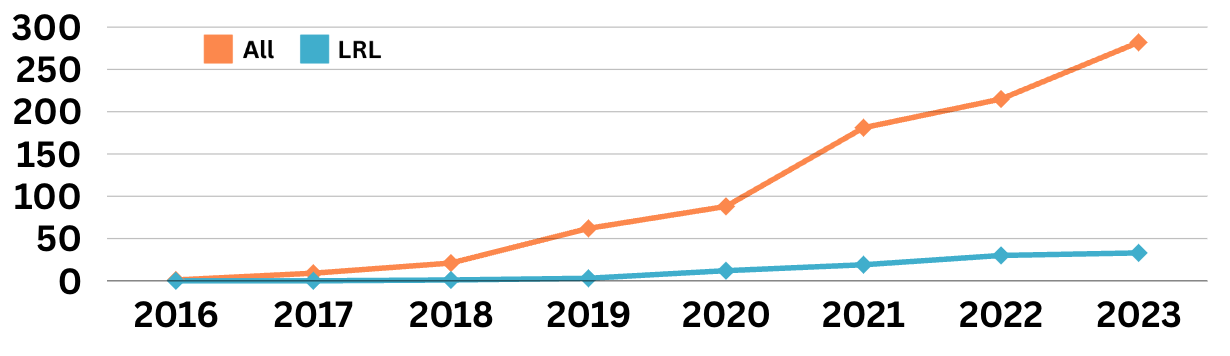}
    \caption{NLP papers on misinformation detection between 2016 and 2023 (\textbf{orange} = \textbf{ALL}; \textbf{blue} = focused on \textbf{low-resource languages (LRLs}))}
    \label{fig:trend}
\end{figure}
Figure \ref{fig:trend} illustrates a significant increase in the number of Natural Language Processing (NLP) papers dedicated to misinformation and fake news detection from 2016 to 2023, with 
83\% of this work focused on monolingual settings for high-resource languages. This focus overlooks the global nature of misinformation, which proliferates across diverse social media platforms where multiple low-resource languages are used simultaneously and often interchangeably. 
The underrepresentation of research in monolingual and multilingual low-resource language settings 
impedes the development of more inclusive and robust detection systems. To bridge this gap, our work provides a comprehensive examination of these aspects, structured around the following research questions:

\noindent \textbf{RQ1}: What are current datasets and research efforts in monolingual and multilingual misinformation detection for low-resource languages?

\noindent \textbf{RQ2}: What challenges have researchers faced engaging in this work and what are future directions?

\begin{figure*}[h!]
    \centering
    \includegraphics[scale=0.43]{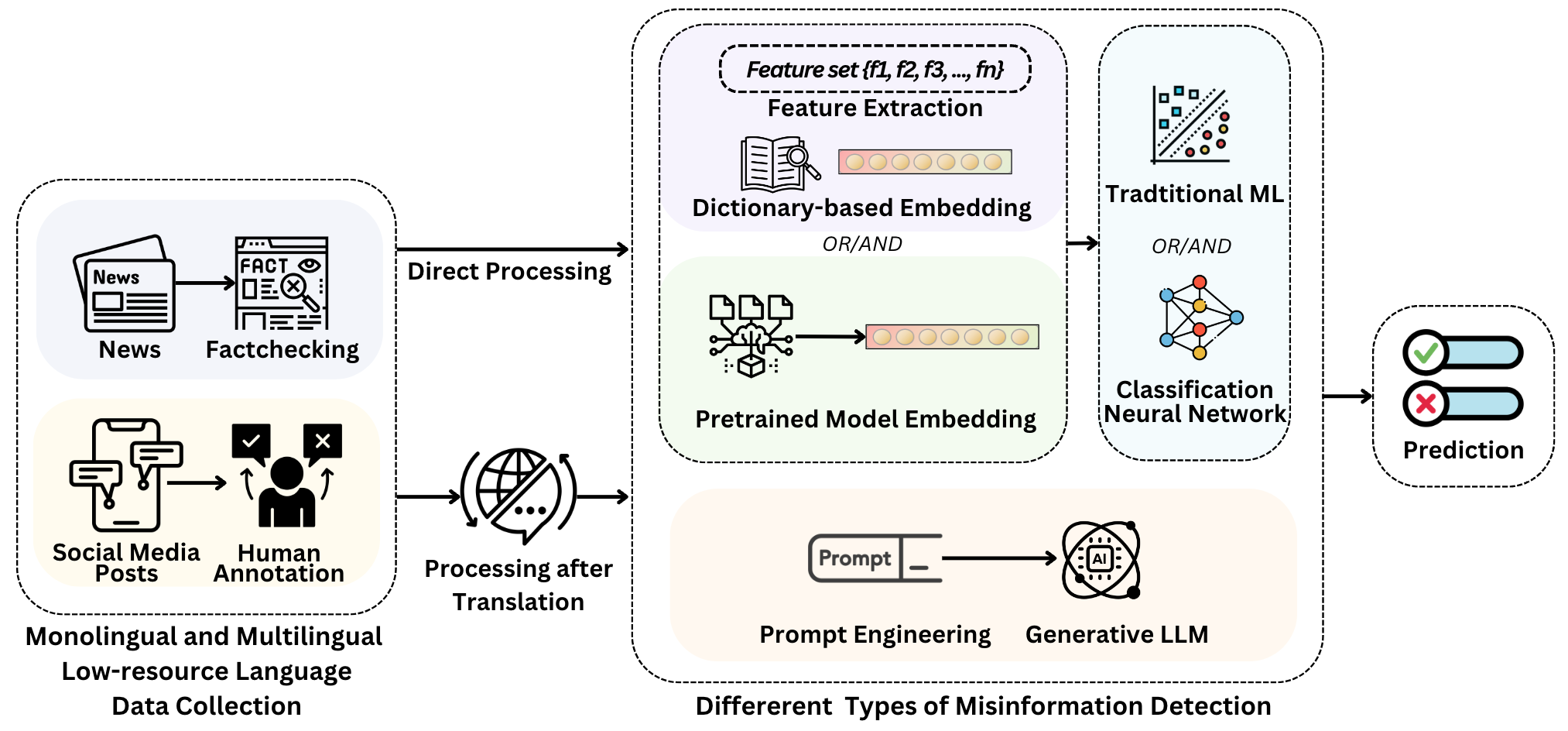}
    \caption{General pipeline of monolingual and multilingual misinformation detection for LRLs }
    \label{fig:pipeline}
\end{figure*}

At a high level, the monolingual and multilingual misinformation detection pipeline for low-resource languages consists of three main phases: (1) data collection and annotation, typically sourced from fact-checked news and human-annotated social media posts; (2) data processing, either direct or indirect, with the latter often involving translation techniques; and (3) detection methods, which generally fall into one of three categories: traditional machine learning classification using feature sets, embedding generation from pre-trained models followed by classification, or leveraging generative large language models (LLMs) through prompt engineering (see Figure \ref{fig:pipeline}). Following, we perform a deeper dive into the structure of this pipeline. In Section \ref{datasets}, we review current monolingual and multilingual datasets for low resource languages.  Section \ref{methods} explores methods employed for these detection tasks, categorized based on the two-layer structure: whether the data is processed directly or indirectly, and the techniques used for processing. In Section \ref{challenges}, we examine the challenges encountered in this field, and Section \ref{future_directions} outlines future research directions.

\section{Background}

\subsection{Low Resource Language Processing in Monolingual and Multilingual Settings}

\noindent \textbf{Definitions.}
In the context of NLP, definitions of \textit{low-resource languages (LRLs)} are far from standardized, as different studies adopt varying definitions based on task-specific needs and data availability rather than a consistent linguistic framework \cite{nigatu2024zeno}. Some studies classify a language as low-resource due to a lack of technical tools and resources \cite{gereme2021combating, devika2024dataset}, while others define it based on sociopolitical and economic constraints \cite{coto2022evaluating, martinez2020cplm}. Broadly speaking, LRLs are characterized by limited resources, such as scarce annotated datasets and digital tools, and they are often less commonly taught with a lower digital presence compared to widely spoken languages \cite{magueresse2020low,singh2008natural}. 
Given these variations, we adopt a clear and consistent definition in this study, scoped around research attention and technological development. We adopt the taxonomy of \citet{joshi-etal-2020-state}, defining high-resource languages as those that have received significant industrial and governmental investment—namely, English, Spanish, German, Japanese, French, Arabic, and Mandarin Chinese—and we refer to all other languages as LRLs, which face varying constraints in terms of available linguistic resources, technological tools, and research infrastructure.

\textit{Multilingual language processing for LRLs} combines linguistics, computer science, and AI to process and analyze natural human language across diverse contexts~\cite{bikel2012multilingual}.

Inside multilingual language processing, \textit{code-switching/code-mixing} represents a special linguistic phenomenon where two or more languages are mixed together within a single utterance, or conversation~\cite{sravani-mamidi-2023-enhancing}. It allows individuals to express unique cultural concepts and connect with or distinguish themselves from others~\cite{toribio2002spanish,ritchie2012social}. 

\noindent \textbf{Current efforts.} 
LRL processing remains a critical concern because of the scarcity of accessible datasets~\cite{shahriar2023question} and the absence of domain knowledge for understanding linguistic structures~\cite{magueresse2020low}. Recent research explores methods to improve NLP performance on LRLs, including but not limited to increasing training through data augmentation~\cite{mahamud2023distributional,chen2023empirical} and fine-tuning domain-specific language models with relatively smaller training corpora~\cite{bhattacharjee2021banglabert,cruz2019evaluating,zhang2023continual}. 

LLMs have also been utilized to process LRLs directly. Large-scale multilingual language models such as XLM-Roberta~\cite{conneau2019unsupervised}, GPT-4~\cite{achiam2023gpt}, and LLama3~\cite{dubey2024llama} utilize a huge amount of independent corpora in different languages during training. This enables them to handle diverse languages with shared vocabulary and text representations. Thus, the capability of cross-lingual transfer are utilized to generalize representations from high-resource languages to low-resource ones.

\subsection{Misinformation and Fake News Detection}

\noindent \textbf{Definitions.} We define \textit{misinformation} as "false or inaccurate information that is deliberately created and is intentionally or unintentionally propagated" \cite{wu2019misinformation}. \textit{Fake news}, a prevalent form of misinformation, is described as "fabricated information that mimics news media content
in form but not in organizational process or
intent" \cite{lazer2018science}. 
In the NLP literature, between 2016 and 2023, the terms misinformation and fake news were often used interchangeably or inconsistently, typically encompassing other categories of misinformation, e.g., rumors, propaganda. For comprehensiveness, we used both terms in our review. 

\noindent \textbf{Current efforts.} 
Existing efforts surveying research on misinformation and fake news detection have concentrated on the ambiguity of these terms, leading to many survey papers establishing taxonomies of misinformation \cite{islam2020deep,hardalov2021survey,aimeur2023fake}. 
Extensive studies have analyzed the topological characteristics of users' interactions, offering a comprehensive view of the linguistic behavior patterns exhibited by social media users \cite{kwak2010twitter,benevenuto2009characterizing}. This work has focused on high-resource languages, particularly English, due to the abundance of resources~\cite{islam2020deep,guo2019future,su2020motivations}. 

In parallel, research in 
communication and related fields has explored the influence of culture and language on the spread and perception of misinformation \cite{aydin2013role}. Yet, these important aspects are seldom highlighted in NLP. 

\section{Inclusion Criteria}

We queried SCOPUS for existing papers on misinformation detection in LRLs. Our specific search query is provided in the Appendix, as is a PRISMA diagram (Figure \ref{fig:prisma}). Our search yielded 188 papers. We excluded papers that were inaccessible, not written in English, or did not approach misinformation detection from a technical perspective. Models designed exclusively for high-resource languages, as defined above, were removed. As multilingual papers do not always mention LRLs even when LRLs are involved, we used multilingual as a keyword and manually filtered results. For dataset review, we identify additional papers by tracing citations related to datasets mentioned in the related work or used for evaluation. This process resulted in 106 dataset and detection-related papers.

\section{Monolingual and Multilingual LRL Datasets for Misinformation Detection}
\label{datasets}
Following, we review existing datasets in our scope, i.e., monolingual and multilingual LRL datasets. Specifically, these datasets were collected through citation tracking from referenced papers. 
A summary is provided in Table~\ref{tab:dataset}.

\begin{table*}[h!]
\centering
\tiny
\begin{tabular}{p{0.25\textwidth}p{0.20\textwidth}p{0.25\textwidth}p{0.05\textwidth}p{0.11\textwidth}}
\toprule
Name & Reference & Language& Size& Source\\
\midrule

Fake.Br&\citet{monteiro2018contributions}&Portuguese&7,200&News/Fact-checking\\
DAST&\citet{lillie2019joint}&Danish&220&Social Media\\
Persian Stance Classification&\citet{zarharan2019persian}&Persian&2,124& Mixed\\
Fake News Filipino &\citet{cruz2019localization}&Filipino&3,206&News/Fact-checking\\
Bend the Truth&\citet{amjad2020bend}&Urdu&900&News/Fact-checking\\
BanFakeNews&\citet{hossain2020banfakenews}&Bangla&49,977&News/Fact-checking\\
FactCorp&\citet{van2020factcorp}&Dutch&1,974&News\\
Slovak Fake News&\citet{sarnovsky2020annotated}&Slovak&1,535&News/Fact-checking\\
Urdu Fake News&\citet{amjad2020urdufake}&Urdu&1,300&News/Fact-checking\\
ProSOUL&\citet{kausar2020prosoul}&Urdu&11,574&News/Fact-checking\\
DanFEVER&\citet{norregaard2021danfever}&Danish&6,407&Wikipedia\\
Kurdish Fake News Dataset&\citet{azad2021fake}&Kurdish&15,000&Social Media\\
TAJ&\citet{samadi2021persian}&Persian&3,720&News/Fact-checking\\
ETH\_FAKE&\citet{gereme2021combating}&Amharic&6,834&Mixed\\
BFNC&\citet{rahman2022fand}&Bengla&5,048&News/Fact-checking\\

Tamil Fake News Dataset&\citet{mirnalinee2022novel}&Tamil&5,273&News/Fact-checking\\
Ax-to-Grind Urdu&\citet{harris2023ax}&Urdu&10,083&News/Fact-checking\\
BanMANI&\citet{kamruzzaman2023banmani}&Bangla&800&Social Media\\
\hline\hline
Fact-checking Dataset&\citet{pvribavn2019machine}&Czech, Polish, Slovak& 24,471&News/Fact-checking\\
FakeCovid & \citet{shahi2020fakecovid} &40 languages&5,182&News/Fact-checking\\
MM-COVID&\citet{li2020toward}&English, Spanish, Portuguese, Hindi, French, and Italian&11,173&News/Fact-checking\\
TALLIP &\citet{de2021transformer}  & English, Hindi, Swahili, Vietnamese, Indonesian&4,900 &News/Fact-checking\\

Indic-covidemic tweet dataset&\citet{kar2021no}&English, Hindi, Bengali&1,438&Social Media\\
X-FACT&\citet{gupta2021x}&25 languages&31,189&News/Fact-checking \\
Fighting the COVID-19 Infodemic Dataset&\citet{shaar2021findings}&Arabic, Bulgarian, and English&9,101&Social Media\\
Deceiver&\citet{vargas2021toward}&Portuguese, English&600&News/Fact-checking\\
MMM&\citet{gupta2022mmm}&Hindi, Bengali, Tamil&10,473&News/Fact-checking\\

MuMiN&\citet{nielsen2022mumin}&41 languages&12,914&Mixed\\
FbMultiLingMisinfo&\citet{barnabo2022fbmultilingmisinfo}&38 Languages&7,334&Social Media\\
Dravidian Languages Fake News Dataset&\citet{malliga2023overview}&Dravidian Languages&5,091&Social Media\\
PolitiKweli&\citet{amol2023politikweli} &Code-mixed Swahili-English, English, Swahili &  29,510&Social Media\\
COVID-19 Vaccine Misinformation Dataset&\citet{kim2023covid}&English, Portuguese, Indonesian&5,952&Social Media\\
DFND&\citet{raja2024fake} &Tamil, Telugu, Kannada, and Malayalam &26,000&Mixed \\
NewsPolyML&\citet{mohtaj2024newspolyml}&English, German, French, Spanish, Italian&32,508&News/Fact-checking\\
\bottomrule
\end{tabular}

\caption{Summary of existing, publicly available datasets for low resource language misinformation detection.}
\label{tab:dataset}
\end{table*}

A key challenge of existing datasets is their narrow focus on high-profile topics \cite{amol2023politikweli,li2020toward}, which limits models' adaptability across different domains. Another issue is inconsistent reporting of annotation procedures. Datasets often come from two sources: 1) human-annotated social media data or 2) news from official or fact-checking websites. Human annotation protocols are often unclear, and fact-checking datasets may lack thorough verification. 

Additionally, datasets created through translation from high-resource languages face challenges with translation accuracy, as automatic systems often fail to preserve semantic and syntactic nuance, leading to misaligned linguistic features that affect dataset quality and downstream task performance.

\section{Approaches for Misinformation Detection in LRLs}
\label{methods}

\begin{figure*}[ht]
    \centering
    \includegraphics[scale=0.43]{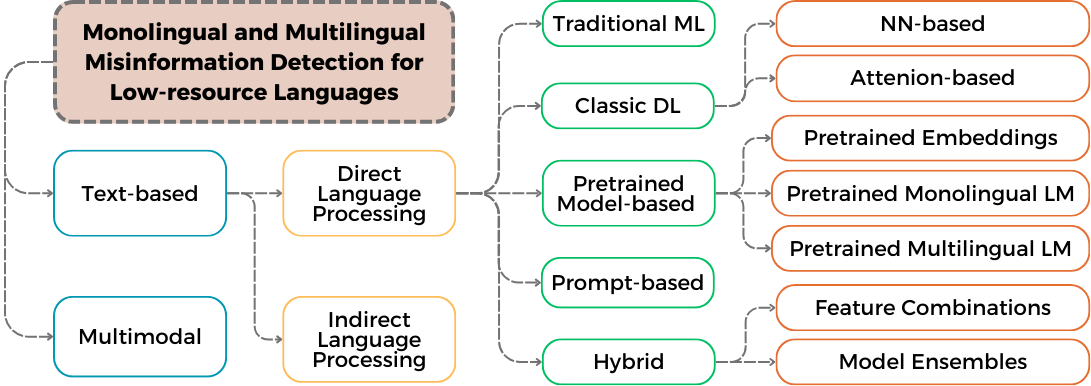}
    \caption{Approaches in monolingual and multilingual misinformation detection for LRLs}
    \label{fig:diagram}
\end{figure*}

Two primary approaches exist for processing monolingual and multilingual LRL data. The majority of research focuses on developing corpora and methods that directly utilize LRL data to address the challenging tasks associated with these languages. However, 
a subset of efforts have been directed towards converting the target language into a better-resourced language for subsequent processing. Following, we categorize models into these two approaches, as shown in 
Figure~\ref{fig:diagram}. We provide examples for each in the main text, 
summarize the full categorization, and include additional details in Tables~\ref{appendix:table1} and~\ref{appendix:table2} in the Appendix.

\subsection{Direct Natural Language Processing}

These approaches directly apply NLP to solve LRL misinformation detection tasks. They can be further categorized into more traditional machine learning, deep learning, and hybrid approaches.

\subsubsection{Traditional Machine Learning} 
Before the advent of transformers \cite{vaswani2017attention}, the dominant approach to misinformation detection involved statistical machine learning (ML). 
~\citet{sharmin2022interaction} deals with Bengali fake news on Facebook as a three-class (fake, real, and satire) classification problem. They test various supervised ML methods and find that XGBoost~\cite{chen2016xgboost} outperforms other classic models.~\citet{coelho2023mucs} addresses fake news detection in code-mixed Malayalam text through the Term Frequency-Inverse Document Frequency (TF-IDF) feature~\cite{beel2017evaluating} and an ensemble model of Multinomial Naive Bayes~\cite{kibriya2005multinomial}, Logistic Regression~\cite{lavalley2008logistic}, and Support Vector Machine (SVM)~\cite{cortes1995support}.

\subsubsection{Classic Deep Learning}
\noindent \textbf{Basic neural network-based methods.} 
~\citet{rafi2022breaking, rasel2022bangla} adapt a Convolutional Neural Network (CNN)-based classication model for fake-news detection in Bengali, fake-news detection in Bangla, and the detection of users that share fake news.~\citet{katariya2022deep} addresses Hindi fake news identification and using a bidirectional long short-term memory (Bi-LSTM) model ~\cite{graves2005bidirectional} which outperforms other baselines.~\citet{mukwevho2024building} focuses on identifying misinformation in LRLs like Tshivenda, where the best performance is obtained through a classification model containing Gated Recurrent Units (GRU)~\cite{cho2014learningphraserepresentationsusing}.

\noindent \textbf{Attention-based methods.}~\citet{hossain2023covtinet} develops CovTiNet, a deep learning model containing an attention module~\cite{bahdanau2014neural} and CNN network for the Covid text identification (CTI) task in Bengali. It achieves 96.61\% accuracy on their dataset, with improved sub-word feature representation and dynamic feature fusion boosting CTI task performance.

\subsubsection{Pre-Trained Models}
Most recently, computational research on misinformation detection has been driven by the impressive performance of 
transformer-based architectures. 

\noindent \textbf{
Pre-trained embeddings.}~\citet{kodali2024bytesizedllm} proposes to use the embedding feature space generated from Subword2vec embeddings and Bi-SLTM architecture~\cite{graves2005bidirectional} for fake news detection in Dravidian languages. 
~\citet{keya2021fake} develops a deep learning model containing the pre-trained GloVe embedding~\cite{pennington2014glove}, Convolutional Neural Network (CNN), and Gated Recurrent Unit (GRU)~\cite{cho2014learningphraserepresentationsusing}. It achieves an accuracy of 98.71\%  for fake news detection in Bangla.

\noindent \textbf{
Monolingual language models.} 
~\citet{amol2023politikweli} focuses on Swahili-English code-switched misinformation detection. Their proposed framework which contains the BERT~\cite{devlin2018bert} model achieves a f1 score of 0.62.~\citet{sultana2023identification} and \citet{kabir2023research} utilize the BERT model in a fake news detection task in Bangla and achieves best performance over BanglaBERT~\cite{bhattacharjee2021banglabert}.~\citet{sivanaiah2023bridging} applies various BERT models for fake news detection in Malay and achieves best performance through MalayBERT~\cite{bendahmane2020phasetransitionsphotonfluid}.

\noindent \textbf{Multilingual language models.} Large language models have been extended to multilingual settings. Pre-training on a huge corpus across many languages, multilingual models enable cross-lingual transfer (a model fine-tuned in one language can be applied to others without further training). To some extent, this makes multilingual models a natural choice for dealing with LRLs.~\citet{chalehchaleh2024multilingual} demonstrates the effectiveness of applying a multilingual language model to non-English fake news detection.~\citet{kim2023covid}, \citet{rahman2022fand}, and \citet{ghayoomi2022deep} also apply XLM-RoBERTa~\cite{conneau2019unsupervised} for COVID-19 vaccine misinformation detection (tweets collected from Brazil, Indonesia, and Nigeria),  fake news classification in Bengli, and COVID‐19 fake news detection in Persian.

\subsubsection{Prompt-Based Methods} 
Research finds that scaling pre-trained models (e.g., scaling model size or data size) often leads to improved model capabilities~\cite{zhao2023survey}. These large language models behave differently from smaller models, particularly on complex tasks; for example, GPT-3 can successfully complete tasks in a few-shot setting by leveraging a small number of examples provided in the prompt. This type of method has been applied to misinformation detection.~\citet{kamruzzaman-etal-2023-banmani} addresses fake news detection and misrepresentation of news in Bengali. They show a fine-tuned version of GPT-3~\cite{brown2020language} outperforms 
chatGPT in zero-shot settings.

\subsubsection{Hybrid Methods}
Others methods combine traditional ML with deep learning approaches. Hybrid methods focus on either feature combinations or model ensembles. 

\noindent \textbf{Feature combinations.}~\cite{kar2021no} focuses on COVID-19 fake news detection in Bengali and Hindi. They extract text embeddings from mBERT with additional features indicating textual and statistical information from Twitter, then identify fake tweets through neural network-based methods.~\citet{salh2023kurdish} focus on Fake News detection in Kurdish through text features (from TF-IDF, count-vector, and word-embedding) and a CNN-based classifier.

\noindent \textbf{Model ensembles.}~\citet{mohawesh2023semantic} develops a new semantic graph attention-based representation learning framework with the multilingual language model to extract structural and semantic representations of texts. This framework outperforms the state-of-the-art techniques for the multilingual fake news detection task (in English, Hindi, Swahili, Vietnamese, and Indonesian).~\citet{sadat2023supervised} addresses the rumor classification in both Bangla and English. The best performance is obtained through a hybrid model containing RNN and Random Forest.

\subsection{Indirect Natural Language Processing}
Indirect natural language processing methods avoid direct analysis of texts in their original languages, instead translating them into languages for which NLP techniques are better developed.~\citet{dementieva2023multiverse} propose a method that summarizes Top-N related articles based on translated titles, integrating new features with BERT-based models for effective fake news detection.

\subsection{Multi-Modal Approaches}
\citet{gupta2022mmm} explores multi-modal misinformation detection in three Indian languages, Hindi, Bengali, and Tamil. Their baseline experiments indicate background knowledge, emotion, and multimodality all influence fake news prediction.

\subsection{Methodological Trends}
\begin{figure}[h!]
    \centering
    \includegraphics[scale=0.5]{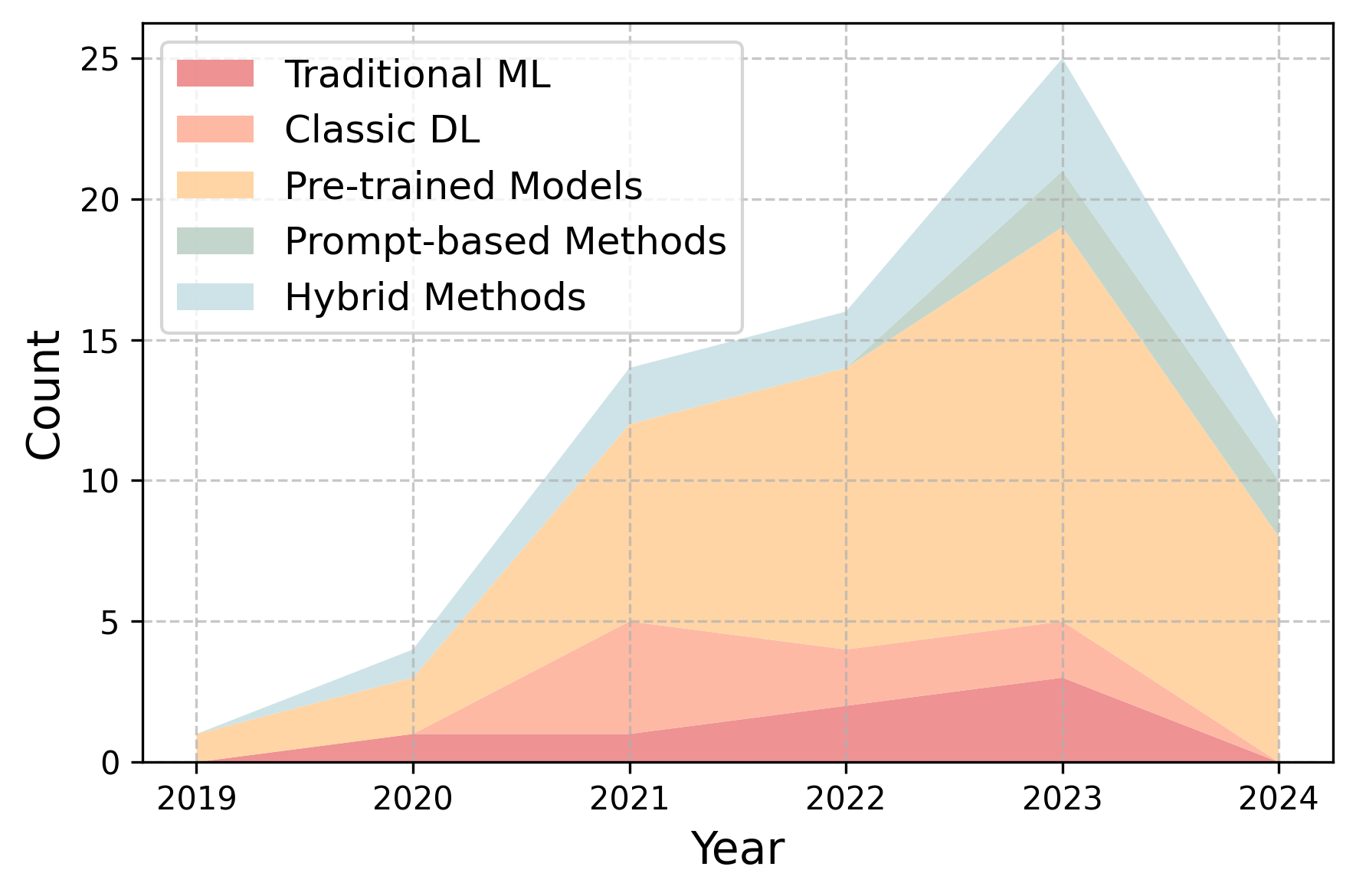}
    \caption{Methodological trends for direct natural language processing }
    \label{fig:model_trend}
\end{figure}

Figure \ref{fig:model_trend} shows the number of papers that utilized the corresponding methods for direct processing of LRLs in each year. 
Our analysis suggests that the research community has moved away from traditional approaches in favor of pre-trained models, which offer improved performance and efficiency in language processing tasks. 
To gain further insight into how these models are applied across different language families, we classified the languages using Ethnologue, a widely recognized resource for language classification \cite{campbell2008ethnologue}. Detailed classifications of the languages studied, grouped by family, are provided in Table~\ref{tab:language_families} in the Appendix.

Figure~\ref{fig:family} shows the distribution of pre-trained models (excluding pre-trained embeddings) across language families, where each language is counted individually. Note that a single paper may cover multiple languages belonging to different families. Our analysis reveals that among LRLs, Indo-European languages have been the primary focus, with the majority of studies leveraging a diverse array of models for detection tasks. In contrast, for families such as Turkic (e.g., Turkish), Kartvelian (e.g., Georgian), and Afro-Asiatic (e.g., Algerian), only a single model has been applied, making comparison and model improvements challenging. Given that there are 143 language families in total \cite{campbell2008ethnologue}, the majority remain understudied.

Prompt-based methods have emerged more recently (2023 and onward), 
driven by the enhanced capabilities of large language models (LLMs), which enable more direct interaction with language data through natural language prompts \cite{white2023prompt,wei2022chain}. 
Although research on LRLs is still emerging, multitask prompted training \cite{sanh2022multitask} and unsupervised cross-lingual representation learning \cite{conneau2019unsupervised} can reduce the need for extensive fine-tuning, which is promising for LRLs with limited training data.

\begin{figure}[h!]
    \centering
    \includegraphics[scale=0.50]{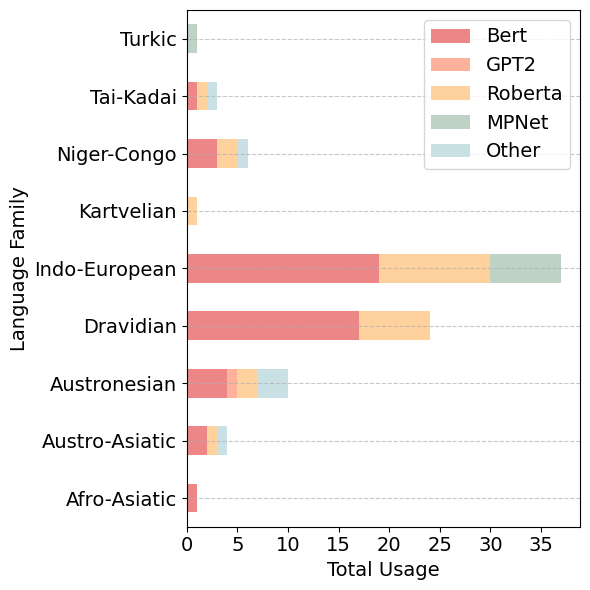}
    \caption{Base pretrained model counts for each language family }
    \label{fig:family}
\end{figure}

\section{Challenges}
\label{challenges}

We have systematically identified and summarized the challenges and limitations faced by researchers in the existing body of literature; synthesized insights from these papers are quantitatively represented in Table~\ref{tab:summary}. Following, we discuss each in further detail.

\begin{table}[ht]
    \centering
    \small
    \setlength{\tabcolsep}{3pt} 
    \renewcommand{\arraystretch}{1.1} 
    \begin{tabular}{p{0.12\textwidth}p{0.35\textwidth}}
        \toprule
        \textbf{Theme} & \makecell{\textbf{Subcategory (Count, Scope)}} \\
        \midrule
        \textbf{Data} & \makecell{Data Scarcity (63, S);\\ Low Data Quality (16, S);\\ Lack of Data Standardization (4, S);\\ Increased Code-switching (4, S)}\\
        \addlinespace
        \textbf{Models} & \makecell{Low Model Generalizability (22, S);\\ Feature Limitations (6, S);\\ Low Model Explainability (3, S);\\ Model Architectural Limitations (3, U);\\ Computational Resource Demands (2, U) }\\
        \addlinespace
        \textbf{Context} & \makecell{Linguistic Heterogeneity (10, S);\\ Dynamic Nature of Fake News (5, U);\\ Absence of Social Context (5, S);\\ Cultural Diversity (3, S) } \\
        \addlinespace
        \textbf{Applications} & \makecell{Lack of Supporting Tools (22, S);\\ Inadequate Real-world Applications (2, S)} \\
        \addlinespace
        \textbf{Research} & \makecell{Low Research Efforts(35, S)} \\
        \bottomrule
    \end{tabular}
    \caption{Challenges in Monolingual and Multilingual LRL Misinformation Detection. Count: number of mentions of the challenge in the reviewed literature; Scope: "S" denotes challenges specific to or amplified in LRL settings, "U" denotes universal challenges.}
    \label{tab:summary}
\end{table}

\subsection{Limited Data Resources}
Data scarcity remains a critical and recurring challenge in the field \cite{agarwal2024deciphering, hariharan2022impact, sharma2024comprehensive}. Beyond scarcity, existing datasets are often affected by issues such as noise, incompleteness, biases, and lack of structure \cite{sivanaiah2023bridging, alghamdi2024fake}, frequently caused by the absence of standardized protocols during the fact-checking process \cite{barnabo2022fbmultilingmisinfo}. Additionally, the increasing prevalence of code-switching further complicates data collection and labeling, requiring specialized approaches to ensure the accuracy and reliability of the datasets \cite{bala2023abhipaw}.

\subsection{Model and Technical Constraints}
In addition to the challenges of computational power demands and architectural limitations common across NLP fields, we also identified the following technical challenges:

\noindent \textit{Limited generalizability of models.} While recent LLMs claim to be multilingual, their performance varies significantly between languages \cite{ignat2023has, rafi2022breaking, huang2023not, ahuja2023mega}. 

\noindent \textit{Lack of external knowledge.} Features, such as metadata \cite{hu2023mr2}, linguistic cues \cite{anirudh2023multilingual}, contextual factors \cite{mohawesh2023semantic}, and social engagement data \cite{li2020toward}, could enhance model performance. 

\noindent \textit{Low explainability.} Lack of models’ transparency is a concern, especially when there is limited knowledge of the LRL in which the fake content is presented \cite{bailer2021challenges}.

\noindent \textit{Insufficient subword representation.} While models are trained on large-scale multilingual corpora, the small vocabulary size for LRLs limits their ability to understand misinformation~\cite{wang-etal-2019-improving}.

\noindent \textit{Low performance of transfer learning.} Cross-lingual transfer has effectively improved downstream task performance for LRLs, but its success is highly dependent on the linguistic similarity between the target languages (LRLs) and those in the training corpus~\cite{philippy2023towards}.

\subsection{Contextual Complexity}
Misinformation's dynamic nature and ambiguous interpretations of irony and satire complicate detection \cite{tufchi2024improved, ernst2024identifying}. In addition, machine translation from LRLs to high-resource languages can reduce accuracy by 23\% in multiclass fake news detection, risking loss of contextual information \cite{saghayan2021exploring}. Thus, constructing corpora and transferring models between languages requires deep cultural understanding to preserve context and language nuances \cite{agarwal2024deciphering}.

\subsection{Inadequate Practical Applications}
Social media platforms focus their moderation efforts on high-resource languages, with their policies often shaped by country-specific regulations \cite{mohawesh2023semantic}. Consequently, valuable research data from the platforms in LRLs are often neglected and underutilized. This imbalance not only limits the practical impact of existing technologies but also perpetuates the vulnerability of LRL communities to misinformation.

\subsection{Insufficient Research Efforts}
LRL misinformation detection is still underexplored, with few studies published to date. This has led to a scarcity of foundational research that could serve as a basis for future work \cite{sultana2023identification}.

\section{Future Directions}
\label{future_directions}
We advocate for increased attention within the NLP research community in several key areas (\textbf{LINGUA}), emphasizing urgency due to globalization of online ecosystems. 

\subsection{(L)anguage-generalizable Models}
One of the most promising avenues for future research is the development of cross-lingual models that can be applied to the diverse online ecosystems where misinformation proliferates. These models have the potential to break down linguistic barriers that currently impede effective misinformation detection \cite{hasanain2023qcri}. Future efforts should focus on refining these models to better capture semantic nuances, while also developing universal embeddings and transfer learning techniques that can be applied across languages without requiring extensive re-training \cite{cruz2019localization}.

\subsection{(IN)terdisciplinary Research in Linguistics and Demography}

Shared semantic structures and regional similarities could be used to group languages for more efficient training and model development \cite{ponti2019modeling}. Linguists could contribute by identifying language usage patterns, while demographers provide insights into cultural and regional contexts, ensuring that models are contextually relevant and adaptable to real-world scenarios. This collaboration would also help to ensure that models are developed with an awareness of ethical concerns, making detection systems more socially responsible \cite{hammerl2022multilingual}. 

\subsection{(G)roundedness in High-Quality Datasets}
Research has demonstrated that multilingual training significantly enhances detection performance across languages, consistently outperforming translation-based approaches \cite{chalehchaleh2024multilingual}. This underscores the importance of improving data availability in LRLs, not only to boost language-specific fake news detection but also to enhance cross-language detection tasks. Establishing rigorous standards for corpus construction and ground-truth labeling in LRLs is essential, as this remains a significant challenge even in monolingual detection tasks \cite{vidgen2020directions, wang2024unappreciated}. Setting such benchmarks will drive progress and ensure that datasets are robust, reliable, and capable of supporting high-performance models across diverse linguistic contexts.

\subsection{(U)nified Multi-Modal Approaches}
While linguistic and cultural variations present challenges in misinformation detection, other data modalities, such as network connections, meme usage, and visual content, may exhibit higher level of uniformity \cite{papantoniou2022deception}. Specifically for LRLs, where text data is scarce, incorporating these modalities can leading to more robust detection systems \cite{chen2025multimodal, tahmasebi2024multimodal, shah2020multimodal}. 

\subsection{(A)pplications for Social Impact}
The impact of misinformation is profound, particularly in regions where LRLs are spoken, and where the effects of misinformation can be exacerbated by limited access to verified information \cite{satapara2024fighting}. As such, there is a need for research that focuses on social good, particularly information integrity. This can be achieved through awarding projects that aim to develop applications tailored to LRLs \cite{siminyu2021ai4d}, as well as through academic and industry partnerships that prioritize ethical AI solutions \cite{mukhija2021designing}. 

\section{Conclusion}
This survey highlights the critical need for further research in monolingual and multilingual misinformation detection for LRLs. By analyzing current datasets, methodologies, and key challenges in the field, we have identified areas requiring immediate attention, such as enhancing data quality, promoting language-generalizable models, and encouraging interdisciplinary collaboration. Insights provided by this review serve as a foundation for future advancements in misinformation detection, ensuring more inclusive and effective systems across diverse cultural and linguistic contexts.

\section{Limitations}
This survey primarily relies on Scopus for the initial collection of relevant papers in the domain. While we extended the review through citation searching within each paper, there remains the possibility that some relevant works were not captured. Additionally, we chose not to focus on developing a comprehensive taxonomy of misinformation, as several existing works already provide detailed taxonomies. Instead, our focus is primarily on the technical aspects of LRL misinformation detection. This narrowed scope may overlook universal dimensions, e.g., sociocultural factors, that may also influence misinformation detection efforts.

\bibliography{main}

\appendix

\section{Appendix}
\label{sec:appendix}

\renewcommand{\thetable}{A\arabic{table}}
\renewcommand{\thefigure}{A\arabic{figure}}
\setcounter{table}{0} 
\setcounter{figure}{0}
\subsection{Search query used in this paper}
\begin{tcolorbox}[boxsep=1pt,left=3pt,right=3pt,top=3pt,bottom=3pt]
Search Query: TITLE-ABS-KEY ( ( "misinformation" OR "fake news" ) AND "detect\*" AND ( "low-resource language\*" OR "code-switch\*" OR "code-mix\*" OR "multilingual" ) ). 

\end{tcolorbox}

\begin{figure*}[ht]
    \centering
    \includegraphics[scale=0.52]{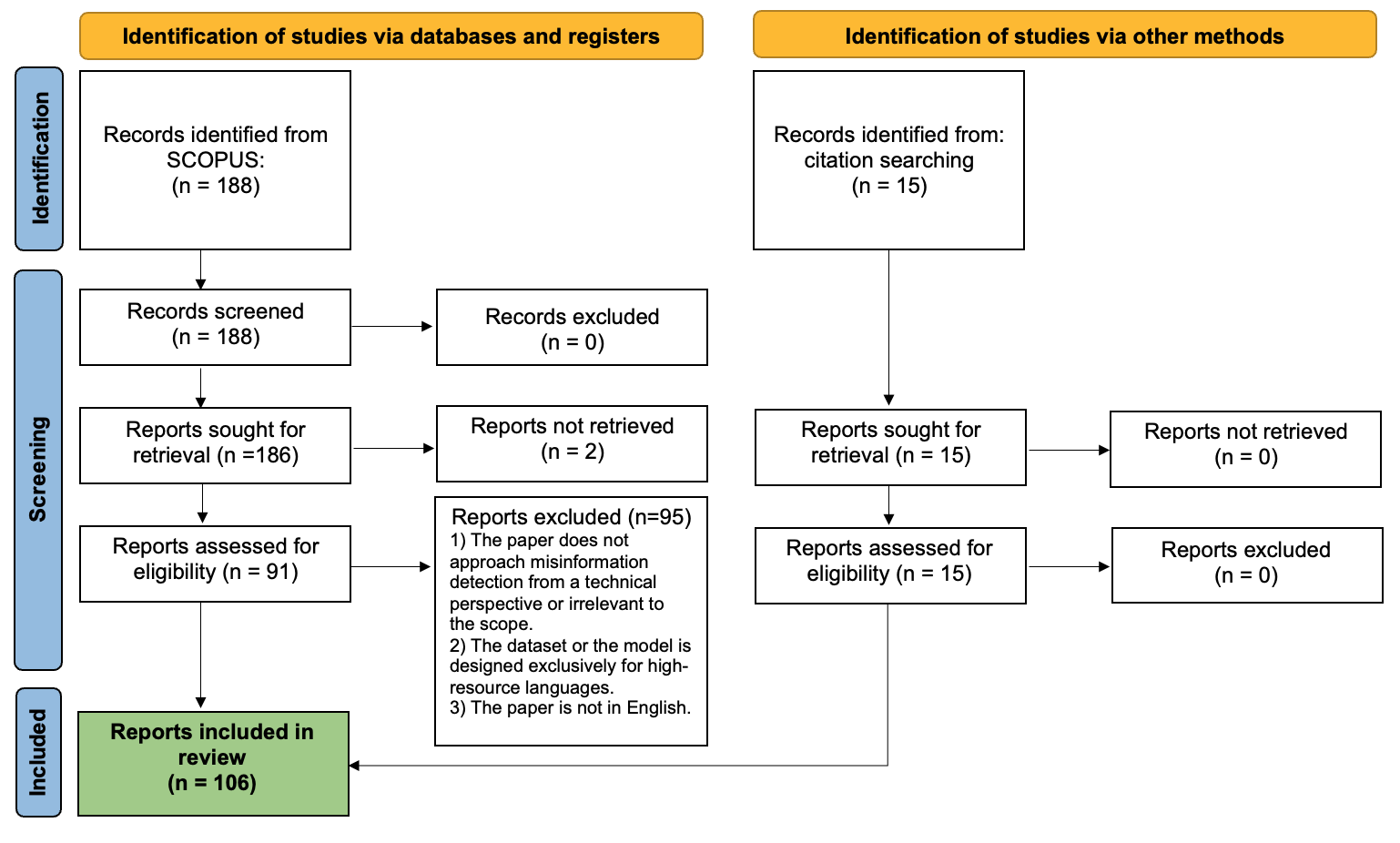}
    \caption{PRISMA diagram for the selection of papers presenting datasets and detection algorithms of misinformation detection}
    \label{fig:prisma}
\end{figure*}

\subsection{Misinformation Detection Paper Categorization}

\begin{table*}[h!]
\centering
\tiny
\begin{tabular}{p{0.20\textwidth}p{0.25\textwidth}p{0.5\textwidth}}
\toprule
\textbf{Category} & \textbf{Papers} & \textbf{Additional Review (not included in the main text)}\\
\midrule
\makecell{Text-based Misinformation Detection->\\ Direct Language Processing->\\ Traditional Machine Learning}&\cite{sharmin2022interaction,coelho2023mucs,kulkarni2023machine,reddy2023optimising,koloski2020multilingual,saghayan2021exploring,amjad2022overview}&
~\citet{kulkarni2023machine} applies grammatical features and traditional ML approaches (e.g., variants of SVM, Adaboost~\cite{hastie2009multi}) to tackle fake news detection in Hindi. They claim that grammatical features with linear SVM achieve the best performance on their dataset.~\citet{reddy2023optimising,koloski2020multilingual} also works on the fake news detection problem through the combination of Term Frequency-Inverse Document Frequency (TF-IDF) and Random Forest~\cite{ho1995random}.\citet{saghayan2021exploring} focuses on fake news detection in Persian tweets related to COVID-19 through n-grams~\cite{shannon1951redundancy} text features and SVM.~\citet{amjad2022overview} 
Optimizes a linear SVM with Stochastic Gradient Descent on fake news
detection in Urdu, indicating that traditional feature-based models outperform contextual representation methods and large neural network algorithms.\\

\makecell{Text-based Misinformation Detection->\\ Direct Language Processing->\\ Classic Deep Learning->\\ Basic Neural Network-based Methods}&~\cite{mohawesh2023multilingual,rafi2022breaking,katariya2022deep,rasel2022bangla,mukwevho2024building,nair2022fake}&~\citet{nair2022fake} proposes a LSTM-based neural network to detect fake news in Malayalam and obtains an accuracy of 93\%.~\citet{mohawesh2023multilingual} proposes a multilingual framework for fake news detection based on a capsule neural network. They prove the effectiveness of this framework in a fake news dataset containing English, Hindi, Swahili, Vietnamese, and Indonesian text.\\

\makecell{Text-based Misinformation Detection->\\ Direct Language Processing->\\ Classic Deep Learning->\\ Attention-based methods}&~\cite{hossain2023covtinet,raja2024adaptive}&~\citet{raja2024adaptive} propose a hybrid deep learning model architecture containing dilated temporal convolutional neural networks (DTCN), bidirectional long-short-term memory (BiLSTM), and a contextualized attention mechanism (CAM) to address the problem of detecting fake news in low-resourced Dravidian languages. Their 
framework achieves an average accuracy of 93.97\% on a fake news dataset in four Dravidian languages.\\

\bottomrule
\end{tabular}

\caption{Categorization of selected papers for traditional machine learning and classic deep learning}
\label{appendix:table1}
\end{table*}

\begin{table*}[h!]
\centering
\tiny
\begin{tabular}{p{0.20\textwidth}p{0.2\textwidth}p{0.55\textwidth}}
\toprule
\textbf{Category} & \textbf{Papers} & \textbf{Additional Review (not included in the main text)}\\
\midrule
\makecell{Text-based Misinformation Detection->\\ Direct Language Processing->\\ Pre-trained Models->\\Incorporate Pre-trained Embeddings}&~\cite{kodali2024bytesizedllm,schwarz2020emet,keya2021fake,gereme2021combating}&~\citet{gereme2021combating} focuses on low-resource African languages, such as Amharic. Authors 
use the Amharic fasttext word embedding (AMFTWE), which performs well on an Amharic fake news detection dataset.~\citet{schwarz2020emet} creates a framework to classify the reliability of short messages posted on social media platforms through text embeddings, called Emet, from a multilingual transformer encoder. 
\\

\makecell{Text-based Misinformation Detection->\\ Direct Language Processing->\\ Pre-trained Models->\\Utilize Monolingual Language Models}&~\cite{amol2023politikweli,sultana2023identification,sivanaiah2023bridging,rasyid2023classification,cruz2019localization,tahir2022ubert22,abdedaiem2023fake,kabir2023research}&~\citet{rasyid2023classification} proposes an Indonesia misinformation detection system with the pre-trained language model and obtains best performance through IndoBERT~\cite{wilie2020indonlu}.~\citet{cruz2019localization} applies GPT-2~\cite{radford2019language} on Filipino fake news detection task the and obtains a classification accuracy of 91\%. ~\citet{tahir2022ubert22} develops a pre-trained BERT model on low resource language Urdu, called  UBERT22. This pre-trained model outperforms Urdu BERT on fake news identification, propaganda classification, and topic categorization tasks.
~\citet{abdedaiem2023fake} proposes a few-shot learning fake news detection model based on sentence transformer fine-tuning,  utilizing no crafted prompts and language model. It highlights that MARBERTv2~\cite{albalawi2023multimodal} stands out as the top-performing model for detecting fake news in the Algerian dialect,  achieving the f1 of 0.71.\\

\makecell{Text-based Misinformation Detection->\\ Direct Language Processing->\\ Pre-trained Models->\\Apply Multilingual Language Models}&~\cite{munir2024bil,araque2023towards,agarwal2024deciphering,tian2023metatroll,hasanain2023qcri,hariharan2022impact,alghamdi2024fake,bala2023abhipaw,raja2024fake,kim2023covid,raja2023nlpt,tabassum2024punny_punctuators,chalehchaleh2024multilingual,ghayoomi2022deep,thaokar2022multi,rahman2022fand,raja2023fake,chaudhari2023empowering,de2021transformer,kazemi2022matching,bhawal2021fake,awal2022muscat,fischer2022identifying,huertas2021countering,kalraa2021ensembling,amjad2020urdufake,panda2021detecting,tarannum2022z,mohtaj2024newspolyml,sivanaiah2022fake}&~\citet{huertas2021countering} presents an approach for detecting misinformation through a semantic-aware multilingual architecture. Their experiments offer promise 
that multilingual models 
may overcome the language bottleneck. ~\citet{munir2024bil} proposes BiL-FaND, an ensemble-based system integrating multiple models designed to analyze distinct aspects of news content in English and Urdu. The framework, containing Multilingual BERT for textual analysis, LSTM models for categorical and numerical data, and a caption-generating model for multimedia content analysis, aims to handle the complexity and nuances of multilingual fake news.~\citet{araque2023towards} applies DistilBERT for vaccine hesitancy detection in English and Italian. Authors find that creating a mixed dataset can support a model capable of classifying instances in two different languages.~\citet{tian2023metatroll} focuses on troll detection (in Thailand, Mexico, and Venezuela) and develops a new model containing a BERT-based feature extractor and an adaptive linear classifier. With the extension of the feature extractor, this framework can be easily adapted to handle multilingual inputs.~\citet{chaudhari2023empowering,de2021transformer,fischer2022identifying,panda2021detecting,sivanaiah2022fake} adapt Multilingual-BERT~\cite{devlin2018bert} on propaganda detection in Hindi, multilingual fake news detection (Hindi, Swahili, Indonesian, Vietnamese), fake news identification in Brazilian Portuguese, COVID-19 misinformation detection (in Bulgarian and Arabic), and fake news detection in low-resource Indian languages (like Tamil, Kannada, Gujarati, and Malayalam), respectively.~\citet{tarannum2022z} focuses on Check-worthiness of tweets (in English, Dutch, and Spanish) and shows that Multilingual-BERT~\cite{devlin2018bert} outperforms other methods on this task.~\citet{mohtaj2024newspolyml} focuses on detecting disinformation in different European languages including English, German, French, Spanish, and Italian. Multilingual-BERT~\cite{devlin2018bert} performs better than Mistral (7B)~\cite{jiang2023mistral} on benchmark tasks in this paper. The benchmark result indicates that the size alone does not dictate a model’s effectiveness for the misinformation detection task. 

~\citet{hasanain2023qcri} addresses the news genre classification task and finds that the classification method containing XLM-RoBERTa~\cite{conneau2019unsupervised} achieves the best performance.~\cite{raja2023fake,hariharan2022impact,raja2023nlpt,tabassum2024punny_punctuators,thaokar2022multi,bala2023abhipaw} concentrate on adapting XLM-RoBERTa~\cite{conneau2019unsupervised} or Multilingual-BERT~\cite{devlin2018bert} to solve the fake news detection in Dravidian languages, such as Telugu, Tamil, and Malayalam.~\citet{kazemi2022matching} focus on automatically finding existing fact-checks for claims made from tweets (in English, Hindi, Portuguese and Spanish). They realize this goal with a classification model containing XLM-RoBERTa~\cite{conneau2019unsupervised} and achieve an accuracy of 86\%.

~\citet{awal2022muscat} proposes Multilingual Source Co-Attention Transformer (MUSCAT), which builds on a multilingual pre-trained language model to perform multilingual rumor detection. ~\citet{raja2024fake,raja2022fake} proposes a novel approach for fake news detection in Dravidian languages based on contextualized word embeddings from a pre-trained language model called MuRIL~\cite{khanuja2021muril}. It also proves its effectiveness in detecting fake news in Urdu~\cite{bhawal2021fake,kalraa2021ensembling,amjad2020urdufake}. To overcome the restriction of sequential input length,~\cite{alghamdi2024fake} proposes a two-stage approach for fake news detection in LRLs: (i) extract only the most relevant content from news through a hybrid extractive and abstraction summarization strategy; (ii) combine multilingual pre-trained model to do further classification.\\

\makecell{Text-based Misinformation Detection->\\ Direct Language Processing->\\ Prompt-based Methods}&~\cite{kamruzzaman-etal-2023-banmani,anirudh2023multilingual,ernst2024identifying,boumber2024blue}&~\citet{ernst2024identifying} explores misinformation detection in multilingual settings and how prompt engineering may improve detection accuracy. To solve the deception detection problem,~\citet{boumber2024blue} proposes a retrieval augmented generation (RAG) system to generate answers through pre-defined prompts. Mistral ~\cite{jiang2023mistral7b}.~\citet{anirudh2023multilingual} focuses on fake news detection in Tamil and affirms the adaptability and effectiveness of BERT and GPT-3.5 models in this context.\\

\makecell{Text-based Misinformation Detection->\\ Direct Language Processing->\\ Hybrid Methods}&~\cite{sharma2023lfwe,kasim2022one,devika2024dataset,saeed2021enriching,tufchi2024improved,kar2021no,salh2023kurdish,kausar2020prosoul,hammouchi2022evidence,rathinapriyaadaptive,mohawesh2023semantic,sadat2023supervised}&\textbf{Hybrid method}: ~\citet{sharma2023lfwe} utilizes word embeddings to extract text features from Hindi news and a traditional machine learning classifier to classify fake news based on these text features. Their best result is obtained through SVM~\cite{cortes1995support}.~\citet{kasim2022one,devika2024dataset} explores the combination of multilingual BERT and traditional machine learning classifiers on fake news (in Urdu and Bengali) and Malayalam fake news.~\cite{saeed2021enriching} detects fake news in Urdu. They propose to use features extracted from the convolutional neural network and n-gram-based features and then apply a traditional majority voting ensemble classifier, utilizing three base models: Adaptive Boosting, Gradient Boosting, and a Multi-Layer Perceptron.

\textbf{Feature combinations}:~\citet{tufchi2024improved} concentrates on the detection of fake news in India. They propose a method to extract text features through Sentence Transformers, Variational Autoencoders, and Topic Modelling and further do fake news detection based on a classic machine learning model. The combination of this feature extraction technique and Random Forest classifier achieves a classification accuracy of 96.80\%.~\citet{kausar2020prosoul} presents a framework to identify propaganda content in the Urdu language. Experiment results show that the combination of News Landscape, word n-gram, and character n-gram features outperform other methods for Urdu text classification.~\citet{hammouchi2022evidence} develops a framework, containing Multilingua-BERT~\cite{devlin2018bert}, for detecting COVID-19 fake news that uses external evidence to verify the veracity of online news in a multilingual setting.~\citet{rathinapriyaadaptive} identifies fake news in Hindi and Tamil through text features (extracted from BERT, transformer networks, and seq2seq network) and a CNN-based classification network. \\

\makecell{Text-based Misinformation Detection->\\ Indirect Language Processing}&~\cite{dementieva2023multiverse}&\\

\makecell{Multi-modal Misinformation Detection}&\cite{gupta2022mmm}&\\

\makecell{Survey Paper}&~\cite{sharma2024comprehensive,amjad2022survey,mahesh2024generalized,liu2024emotion,akter2023deep,touahri2024survey,busioc2020literature,bailer2021challenges,kumar2021introduction}&\\

\bottomrule
\end{tabular}

\caption{Categorization of selected papers for advanced methods}
\label{appendix:table2}
\end{table*}

\begin{table*}[h]
    \centering
    \begin{tabular}{lll}
        \hline
        \textbf{Language Family}& \textbf{Languages Studied}  \\
        \hline
        Afro-Asiatic & Algerian\\
        Turkic & Turkish\\
        Austro-Asiatic & Vietnamese\\
        Austronesian & Filipino, Indonesian, Malay\\
        Dravidian & Gujarati, Kannada, Malayalam, Tamil, Telugu \\
        Indo-European & \makecell{Bengali, Portuguese, Bulgarian, Czech, Dutch, Greek,\\ Hindi, Italian,Marathi, Persian, Polish, Russian, Urdu} \\
        Kartvelian & Georgian\\
        Niger-Congo & Nigeria, Swahili \\
        Tai-Kadai & Thai\\
        \hline
    \end{tabular}
    \caption{Unique Languages per Language Family}
    \label{tab:language_families}
\end{table*}

\end{document}